\pdfoutput=1
\documentclass[11pt]{article}

\usepackage[final]{acl}

\usepackage{times}
\usepackage{latexsym}
\usepackage[T1]{fontenc}
\usepackage[utf8]{inputenc}
\usepackage{microtype}
\usepackage{inconsolata}
\usepackage{graphicx}
\usepackage{booktabs}
\usepackage{multirow}
\usepackage{amsmath,amssymb,amsthm}
\usepackage{array}
\usepackage{xspace}
\usepackage{xcolor}

\newcommand{\method}{RebaseKV\xspace}
\newcommand{\ttft}{TTFT\xspace}
\newcommand{\tprefill}{t_{\mathrm{prefill}}}
\newcommand{\tdecode}{t_{\mathrm{decode}}}
\newcommand{\tboot}{t_{\mathrm{boot}}}
\newcommand{\kh}{K_{\mathrm{hide}}}
\newcommand{\cache}{\mathrm{KV}}
\newcommand{\KL}{D_{\mathrm{KL}}}

\newtheorem{proposition}{Proposition}
\newtheorem{theorem}{Theorem}
\newtheorem{corollary}{Corollary}

\title{RebaseKV: Hiding Prefill Latency in Multi-Agent LLMs via Optimistic Cache Alignment}

\author{
  \textbf{Jiayu Qin\textsuperscript{1}}\thanks{Work done during an internship at Amazon.},
  \textbf{Ankith M S\textsuperscript{2}},
  \textbf{Harpreet Kaur\textsuperscript{2}},
  \textbf{Kevin Zhai\textsuperscript{3}}\footnotemark[1],
  \textbf{Sumit Negi\textsuperscript{2}},
  \textbf{Changyou Chen\textsuperscript{1}}
  \\
  \textsuperscript{1}State University of New York at Buffalo\quad
  \textsuperscript{2}Amazon\quad
  \textsuperscript{3}University of Central Florida
}

\begin{document}
\maketitle

\begin{abstract}
In multi-agent LLM systems, agent-specific role instructions shift a largely shared context and invalidate exact prefix caching, so every handoff places a full prefill on the time-to-first-token (TTFT) critical path. We introduce RebaseKV, a training-free serving method based on Optimistic Cache Alignment. At an agent transition, RebaseKV decodes from a position-adjusted cache inherited from the preceding agent while the exact cache for the new prompt is built concurrently on the same GPU; after a short optimistic window, it replays the drafted tokens on the fresh cache and continues standard decoding, so the inherited state only hides latency. With the exact prefill running concurrently, RebaseKV delivers the first token of each handoff 2.95x faster than full recomputation in a five-agent GSM8K pipeline (46 vs.\ 135 ms) and 2.54x faster with Qwen3-8B, with no stall and negligible end-to-end overhead. Routing format-fixing agents, such as an answer-first decision maker, to the exact path makes accuracy statistically indistinguishable from recomputation while the other handoffs keep the low TTFT. We further prove that stale-state divergence is confined to the optimistic window, bound the prefix divergence in logit space, and give a certificate for greedy-token agreement.
\end{abstract}

\section{Introduction}

Multi-agent large language model (LLM) applications solve a task through a sequence of model calls: one agent proposes a solution, another critiques it, a specialist repairs it, and a final agent decides what to return. These calls repeatedly include the same user request, shared constraints, intermediate outputs, and workflow state. Their prompts are therefore highly redundant.

That redundancy does not translate directly into standard key--value (KV) cache hits. Production inference engines reuse cached attention states when the next request has an identical prefix. In a multi-agent workflow, however, each role typically begins with a different system instruction. Changing that role prefix shifts the shared tokens and changes the context under which their hidden states were computed. Exact prefix caching misses, even when most of the text is repeated, and each agent handoff can expose a full prompt prefill before the first token is emitted.

A tempting alternative is to decode directly from the preceding agent's cache. This gives low latency, but the cache remains conditioned on the wrong role. In our GSM8K pipeline, permanent stale reuse reduces accuracy from 90.6\% to 75.4\%. More diagnostically, a Programming-Expert agent emits code in only 3\% of cases when it inherits the Analyst's role-conditioned cache, compared with 100\% under full recomputation. The central challenge is therefore not merely positional alignment: the serving system must obtain a fast first token without letting a role-misaligned state govern the entire answer.

We propose \method, a training-free serving architecture that changes the purpose of approximate cache reuse. Rather than correcting an inherited cache until it is accurate enough for full decoding, \method uses it as a temporary foreground state. While a short optimistic prefix is generated, an exact cache for the new agent prompt is constructed on a background stream. The drafted prefix is then replayed against that fresh cache, after which decoding proceeds from a prompt-consistent state. Exact prefill is moved off the first-token critical path, while stale-state exposure is confined to a small number of accepted prefix tokens.

This execution pattern exploits the asymmetry between parallel, compute-intensive prefill and sequential, often memory-bandwidth-bound decoding \citep{agrawal2023sarathi,agrawal2024sarathiserve}. It is complementary to throughput-oriented prefill--decode disaggregation: \method overlaps the two phases within one agent transition to optimize interactive responsiveness, at negligible end-to-end cost.

Our contributions are:
\begin{itemize}
    \item We identify role-prefix divergence as a concrete source of repeated prefill and behavioral failure in multi-agent LLM pipelines.
    \item We introduce Optimistic Cache Alignment, which overlaps an inherited-cache draft with exact prompt prefill and then rebases the accepted prefix onto the fresh cache, together with the execution requirements (stream priority) and an agent-routing rule that make it work in practice.
    \item We prove that stale-state divergence is confined to the first $K$ optimistic steps, derive a logit-space KL bound, and give a sufficient condition for greedy-prefix identity that is checkable at no extra cost.
    \item We measure the method with the exact prefill genuinely running concurrently on one GPU on two model families, show that an agent-routing rule makes accuracy match recomputation, and compare against RelayCaching.
\end{itemize}
Code, per-question outputs, and the two-stream driver are available at \url{https://github.com/jiayujack/rebasekv}.

\section{Related Work}

\paragraph{Exact and modular cache reuse.}
PagedAttention and radix-tree serving improve KV memory management and exact prefix reuse \citep{kwon2023vllm,zheng2024sglang}. Prompt Cache exposes reusable prompt modules \citep{gim2024promptcache}, while workflow-aware systems such as KVFlow use predictable agent execution to guide eviction and prefetching \citep{wu2025kvflow}. These methods are most effective when cached states remain valid for the new prefix. RebaseKV targets shared text whose position and preceding role context have changed.

\paragraph{Non-prefix and cross-context reuse.}
CacheBlend selectively recomputes fused knowledge chunks, EPIC introduces position-independent caching with boundary correction, and KVLink combines positional adjustment with learned link tokens \citep{yao2025cacheblend,hu2025epic,yang2025kvlink}. DroidSpeak identifies critical layers for cross-model cache reuse \citep{liu2025droidspeak}, and RelayCaching rectifies cache divergence across collaboration stages \citep{geng2026relaycaching}. These systems repair or recompute reused states before relying on them, so the repair stays on the TTFT path and the repaired cache becomes the long-lived state. RebaseKV instead accepts a bounded optimistic prefix and makes the exact prompt-conditioned cache the long-lived state; Section~\ref{sec:relay} compares the two directly.

\paragraph{Multi-agent serving.}
KVCOMM estimates cache deviations online using an anchor pool for cross-context alignment \citep{ye2025kvcomm}. TokenDance exploits all-gather structure for collective cache sharing, while TokenCake and Continuum coordinate cache residency with agent or tool-call schedules \citep{chen2025tokendance,wu2025tokencake,luo2025continuum}. RebaseKV is orthogonal to these memory and scheduling policies: it addresses the latency of one prefix-divergent handoff.

\paragraph{Prefill--decode co-scheduling.}
DistServe disaggregates prefill and decode across GPU pools, and Sarathi-Serve chunks prefills to reduce decode stalls \citep{zhong2024distserve,agrawal2024sarathiserve}. RebaseKV uses the same resource asymmetry at a finer granularity, overlapping exact prefill with the first few decode steps of the next agent.

\section{RebaseKV}
\label{sec:method}

\begin{figure*}[t]
    \centering
    \includegraphics[width=\textwidth]{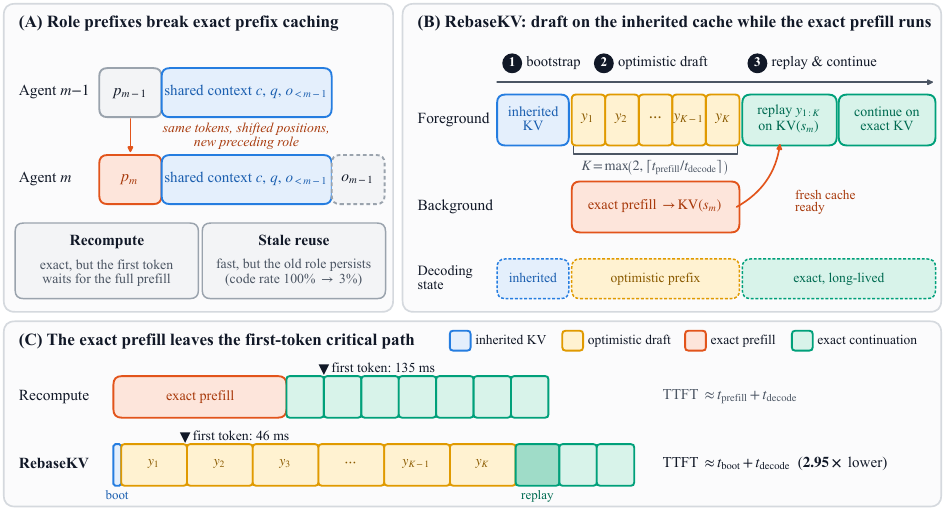}
    \caption{Optimistic Cache Alignment. (A) Consecutive agents share most of their prompt but differ in role prefixes, so exact prefix caching misses: recomputation blocks on prefill, and stale reuse keeps the previous role. (B) RebaseKV drafts $K$ tokens from the inherited cache while the exact $\cache(s_m)$ is built in the background, replays the draft on the fresh cache, and continues on exact KV. (C) The exact prefill leaves the first-token critical path (measured medians from Table~\ref{tab:twostream}).}
    \label{fig:overview}
\end{figure*}

\subsection{Setting and Failure Mode}

Consider an $M$-agent pipeline. Agent $m$ receives
\begin{equation}
    s_m = [p_m; c; q; o_{<m}],
    \label{eq:prompt}
\end{equation}
where $p_m$ is its role or system prompt, $c$ contains shared instructions, $q$ is the user request, and $o_{<m}$ contains previous agents' outputs. Consecutive prompts share most of $c$, $q$, and $o_{<m}$, but replacing $p_{m-1}$ with $p_m$ shifts shared tokens and changes their prefix context.

Full recomputation builds $\cache(s_m)$ before decoding and has approximate first-token latency
\begin{equation}
    \mathrm{TTFT}_{\mathrm{recomp}} \approx \tprefill + \tdecode.
\end{equation}
Permanent stale reuse removes $\tprefill$ from this path, but it decodes the response from a cache conditioned on $p_{m-1}$. The role-following failure in Section~\ref{sec:format} shows why positional correction alone is insufficient.

\subsection{Optimistic Cache Alignment}

For every rebased agent, RebaseKV executes three stages (Figure~\ref{fig:overview}).

\paragraph{1. Bootstrap an inherited cache.}
A greedy $n$-gram matcher identifies spans shared with the preceding prompt. Novel tokens are computed normally; matched KV states are reused. If a matched span moves by $\Delta$ positions in a model with rotary position embeddings, its keys are shifted as
\begin{equation}
    \hat{k}^{(\ell)}_n = R_{\Delta} k^{(\ell),\mathrm{src}}_{n-\Delta},
    \label{eq:rope}
\end{equation}
where $R_{\Delta}$ applies the relative RoPE rotation \citep{su2021roformer}. This corrects the positional phase, while deep-layer states still reflect the preceding role, so RebaseKV uses the result only as a temporary draft cache.

\paragraph{2. Draft while exact prefill runs.}
The foreground stream emits $K$ optimistic tokens from the inherited cache. Concurrently, a background stream computes the exact $\cache(s_m)$ under the new role. The smallest draft window expected to cover this work is
\begin{equation}
    \kh = \left\lceil \frac{\tprefill}{\tdecode} \right\rceil,
    \qquad K = \max(2,\kh),
    \label{eq:khide}
\end{equation}
using recent timing measurements for the current prompt and device. With successful overlap,
\begin{equation}
    \mathrm{TTFT}_{\mathrm{RebaseKV}} \approx \tboot + \tdecode,
\end{equation}
so exact prefill no longer blocks the first token.

\paragraph{3. Replay and continue.}
When the fresh cache is ready, the $K$ accepted draft tokens are replayed in one forward pass on top of $\cache(s_m)$. This creates an exact cache for the new prompt conditioned on the accepted optimistic prefix. All subsequent tokens are decoded from this fresh state. Stale KV tensors therefore never propagate beyond the replay boundary (Appendix~\ref{app:timeline} illustrates the schedule).

\subsection{Executing the Window on One GPU}
\label{sec:execution}

RebaseKV runs the draft on a high-priority CUDA stream and launches the exact prefill from a separate scheduler thread, as the scheduler of an inference engine does, so the first draft step never queues behind prefill kernels; this stream priority is an essential part of the design (Section~\ref{sec:twostream}). The window is either fixed by Equation~\ref{eq:khide} from the uncontended prefill time or \emph{event-driven}, drafting until the background stream signals that the fresh cache is ready. Because contended draft steps slow down together with the prefill, a fixed window covers the contended prefill in every measured handoff. The two caches coexist only during the window; for Qwen2.5-7B (28 layers, 4 KV heads, head dimension 128, bf16) the extra cache occupies 56 KB per token, about 45 MB for an 800-token prompt.

\subsection{Agent Routing and the Exact Path}
\label{sec:routing}

The optimistic prefix sets how a response begins. RebaseKV therefore routes \emph{format-fixing} agents, whose first tokens determine a structure (a tool-call or JSON header, a single-letter decision, an answer-first summary, final code synthesis), to the exact path and rebases free-form reasoning and critique agents. The exact path still hides most of its prefill: the known part of the agent's prompt can be prefilled while the preceding agent is still decoding (WarmupKV, Appendix~\ref{app:ablations}), leaving only the preceding agent's output on the critical path, so the two mechanisms hide prefill in the previous handoff and in the current one, respectively.

The routing decision uses information the system already has: shared-span coverage from the matcher, $\kh$ from recent timing (falling back to the exact path when it exceeds a small budget), and the agent type from the workflow configuration. A draft-agreement monitor provides an online signal at no extra cost. The fresh-cache logits at every draft position are computed anyway, position 1 by the exact prefill and positions $2,\dots,K$ by the replay forward, so the agreement between drafted and fresh top-1 tokens, and the certificate of Proposition~\ref{prop:greedy}, are available after every handoff to move an agent between the two paths. In our MMLU and HumanEval pipelines the final agent (a single-letter decision and final code synthesis, respectively) uses the exact path; for GSM8K we report both the all-optimistic configuration and the configuration with the Decision Maker on the exact path.

\section{Theoretical Characterization}
\label{sec:theory}

We formalize the distinction between permanent stale reuse and RebaseKV; proofs are in Appendix~\ref{app:proofs}. The results show that replay stops all \emph{stale-state} mismatch after the optimistic prefix.

\subsection{Notation and Permanent Stale Exposure}

Fix an agent prompt $s_m$ and a generation horizon $T$. Let $P_{1:T}$ denote the full-recompute sequence distribution and $Q_{1:T}$ the distribution produced by decoding entirely from a stale cache. Their autoregressive factorizations are
\begin{equation}
\begin{aligned}
P_{1:T}(y_{1:T})
&=\prod_{t=1}^{T}p_t(y_t\mid y_{<t}),\\
Q_{1:T}(y_{1:T})
&=\prod_{t=1}^{T}q_t(y_t\mid y_{<t}),
\end{aligned}
\label{eq:pq-factorization}
\end{equation}
where $p_t$ is the exact fresh-cache next-token distribution and $q_t$ is the stale-cache distribution under the same history. We suppress their dependence on $s_m$.

For a history $h$, define the conditional mismatch
\begin{equation}
d_t(h)=\KL\!\left(p_t(\cdot\mid h)\|q_t(\cdot\mid h)\right).
\label{eq:conditional-mismatch}
\end{equation}
The KL chain rule gives
\begin{equation}
\KL(P_{1:T}\|Q_{1:T})
=\sum_{t=1}^{T}\mathbb{E}_{y_{<t}\sim P}[d_t(y_{<t})].
\label{eq:stale-chain}
\end{equation}
Define the nonnegative one-step term
\begin{equation}
\delta_t=\mathbb{E}_{y_{<t}\sim P}[d_t(y_{<t})].
\label{eq:delta-def}
\end{equation}
Then $\Delta_T^{\mathrm{stale}}=\sum_{t=1}^{T}\delta_t$ measures cumulative stale exposure.

\begin{proposition}[Cumulative stale exposure]
\label{prop:cumulative-stale}
For permanent stale-cache decoding,
\begin{equation}
\Delta_{T+1}^{\mathrm{stale}}
=\Delta_T^{\mathrm{stale}}+\delta_{T+1}
\ge \Delta_T^{\mathrm{stale}}.
\label{eq:cumulative-stale}
\end{equation}
\end{proposition}

Thus, even a modest per-token mismatch can continue to accumulate when a role-misaligned cache governs the complete response. This is the behavior that RebaseKV's replay boundary is designed to stop.

\subsection{Finite-Window Error Localization}

Let $R^{(K)}_{1:T}$ denote the RebaseKV sequence distribution. During the first $K$ steps, it uses the stale foreground kernel. After the exact prompt cache is ready, the accepted prefix $y_{1:K}$ is replayed on that cache, and all later tokens use the exact kernel conditioned on the accepted history:
\begin{equation}
\begin{aligned}
R^{(K)}_{1:T}(y_{1:T})
&=\left[\prod_{t=1}^{K}q_t(y_t\mid y_{<t})\right]\\[-2pt]
&\quad\times
\left[\prod_{t=K+1}^{T}p_t(y_t\mid y_{<t})\right].
\end{aligned}
\label{eq:rebase-factorization}
\end{equation}

\begin{theorem}[Finite-window error localization]
\label{thm:finite-window}
For any $T\ge K$,
\begin{equation}
\begin{aligned}
\KL\!\left(P_{1:T}\|R^{(K)}_{1:T}\right)
&=\KL\!\left(P_{1:K}\|Q_{1:K}\right)\\
&=\sum_{t=1}^{K}\delta_t.
\end{aligned}
\label{eq:finite-window}
\end{equation}
\end{theorem}

The reverse-direction identity also reduces to the prefix:
\begin{equation}
\KL\!\left(R^{(K)}_{1:T}\|P_{1:T}\right)
=\KL\!\left(Q_{1:K}\|P_{1:K}\right),
\label{eq:reverse-finite-window}
\end{equation}
where expectations are taken under the RebaseKV prefix distribution.

\begin{corollary}[Continuation divergence is prefix-bounded]
\label{cor:continuation}
Let $P_{K+1:T}$ and $R^{(K)}_{K+1:T}$ be the marginal continuation distributions. Then
\begin{equation}
\KL\!\left(P_{K+1:T}\|R^{(K)}_{K+1:T}\right)
\le
\KL\!\left(P_{1:K}\|Q_{1:K}\right).
\label{eq:continuation-bound}
\end{equation}
\end{corollary}

Once the optimistic prefix is fixed, the long-lived cache is exact for $[s_m;y_{1:K}]$ and no further stale-cache terms are introduced, so Corollary~\ref{cor:continuation} predicts that continuation mismatch plateaus after replay instead of growing with the response horizon.

\subsection{Bounding the Optimistic-Prefix KL}

At step $t$, let
\begin{equation}
p_t=\operatorname{softmax}(z_t/\tau),\qquad
q_t=\operatorname{softmax}(\tilde z_t/\tau),
\label{eq:softmax-kernels}
\end{equation}
where $z_t$ and $\tilde z_t$ are the fresh- and stale-cache logits under the same history, and $\tau>0$ is temperature. Define $\Delta_t=\tilde z_t-z_t$ and
\begin{equation}
\operatorname{span}(\Delta_t)
=\max_v\Delta_t(v)-\min_v\Delta_t(v).
\label{eq:logit-span}
\end{equation}
For compactness, let
\begin{equation}
b_t=\mathbb{E}_{y_{<t}\sim P}\!\left[
\operatorname{span}(\tilde z_t-z_t)^2\right].
\label{eq:bt-def}
\end{equation}

\begin{proposition}[Optimistic-prefix KL bound]
\label{prop:logit-bound}
For a draft of length $K$,
\begin{equation}
\KL\!\left(P_{1:T}\|R^{(K)}_{1:T}\right)
\le \frac{1}{8\tau^2}\sum_{t=1}^{K}b_t.
\label{eq:logit-bound}
\end{equation}
If $\operatorname{span}(\tilde z_t-z_t)\le\rho_t$ for every $t\le K$, then
\begin{equation}
\KL\!\left(P_{1:T}\|R^{(K)}_{1:T}\right)
\le \frac{1}{8\tau^2}\sum_{t=1}^{K}\rho_t^2.
\label{eq:rho-bound}
\end{equation}
For a uniform $\rho_t\le\rho$, this becomes $K\rho^2/(8\tau^2)$.
\end{proposition}

The bound makes the latency--risk interaction explicit: exposure grows at most linearly with $K$ under a uniform logit-span bound, and quadratically with the magnitude of the fresh--stale logit perturbation. It also gives a principled target for online window selection.

\subsection{Greedy-Decoding Certificate}

Let $a_t^\star=\arg\max_a z_t(a)$ be the fresh-cache top-1 token, with margin
\begin{equation}
\gamma_t=z_t(a_t^\star)-\max_{a\ne a_t^\star}z_t(a).
\label{eq:fresh-margin}
\end{equation}

\begin{proposition}[Greedy prefix certificate]
\label{prop:greedy}
Under greedy decoding, if
\begin{equation}
\|\tilde z_t-z_t\|_\infty<\gamma_t/2
\label{eq:greedy-certificate}
\end{equation}
for every optimistic step $t\le K$, then the stale-cache token equals the full-recompute token at each such step. If the condition holds along the complete optimistic trajectory, the drafted prefix is identical to full recomputation, and replay yields the same subsequent greedy continuation up to numerical roundoff.
\end{proposition}

The certificate is checkable after every handoff at no extra cost, because the exact prefill and the replay produce the fresh logits at every draft position (Section~\ref{sec:routing}).

\subsection{Choosing the Draft Window}

Equation~\ref{eq:khide} gives the number of decode steps needed to cover exact prefill. A divergence budget $\epsilon$ and Proposition~\ref{prop:logit-bound} define
\begin{equation}
K_{\epsilon}=\max\left\{k:
\frac{1}{8\tau^2}\sum_{t=1}^{k}\rho_t^2\le\epsilon
\right\}.
\label{eq:kepsilon}
\end{equation}
If $K_{\mathrm{hide}}\le K_{\epsilon}$, choosing the shortest latency-hiding window is sufficient under the bound. Otherwise the system rebases earlier or routes the agent to the exact path. RebaseKV obtains $K_{\mathrm{hide}}$ from runtime measurements or ends the window on the fresh-cache event (Section~\ref{sec:execution}).

\section{Experimental Setup}
\label{sec:setup}

\paragraph{Models and tasks.}
We evaluate Qwen2.5-7B-Instruct on GSM8K and MMLU, and Qwen2.5-Coder-7B-Instruct on HumanEval \citep{qwen2025qwen25,cobbe2021gsm8k,hendrycks2021mmlu,chen2021humaneval}. The evaluated sets contain 1,319 GSM8K questions (1,318 completed by every method), 153 MMLU questions, and 161 HumanEval problems. As a second model family we run Qwen3-8B with thinking disabled on the first 300 GSM8K questions. All pipelines are fully connected and follow the KVCOMM-style role prompts \citep{ye2025kvcomm}: GSM8K uses five agents (Math Solver, Analyst, Programming Expert, Inspector, Decision Maker), and MMLU and HumanEval use five worker agents followed by a final answer agent. Routing follows Section~\ref{sec:routing}.

\paragraph{Accuracy protocol.}
For each question, we draw four final answers: one greedy sample (the timed one) and three samples at temperature $0.7$. We report their mean correctness as \emph{avg@4} and the greedy accuracy separately; differences between methods are paired by question and given with 95\% bootstrap intervals (10,000 resamples). We also use a GSM8K role probe that checks whether the Programming-Expert agent emits a code block. The accuracy harness builds the inherited cache from the preceding agent's complete prompt, so the draft starts without seeing the new role prompt, the hardest setting for the optimistic window.

\paragraph{Latency protocols.}
\emph{Harness TTFT} (Table~\ref{tab:main}; dedicated NVIDIA A100 40GB) is the per-agent median foreground first-token latency in the accuracy harness, where the exact prefill of every agent is executed and timed to set $K$. \emph{Two-stream execution} (Section~\ref{sec:twostream}; one RTX A6000 48GB) runs the exact prefill concurrently on a second CUDA stream and event-times the foreground draft, so TTFT includes contention; every question runs a complete Recompute pipeline and a complete RebaseKV pipeline.

\paragraph{Baselines.}
\emph{Recompute} prefills every agent prompt exactly. \emph{Stale} uses inherited states for the entire response. \emph{RebaseKV} uses the method in Section~\ref{sec:method}. We also report KVCOMM as an online cache-alignment baseline where available, WarmupKV as an exact pipelined alternative, and RelayCaching with its public implementation (Section~\ref{sec:relay}).

\section{Results}
\label{sec:results}

\begin{table}[t]
\centering
\small
\setlength{\tabcolsep}{3.6pt}
\begin{tabular}{llrrr}
\toprule
Data & Method & avg@4 (\%) & TTFT (ms) & $K$ \\
\midrule
\multirow{4}{*}{GSM8K}
 & Recompute & $90.6{\pm}0.7$ & $79.3{\pm}7.2$ & -- \\
 & Stale     & $75.4{\pm}0.8$ & $26.2{\pm}3.1$ & 0 \\
 & \method   & $88.0{\pm}0.7$ & $26.3{\pm}0.2$ & 3.7 \\
 & KVCOMM    & $83.1{\pm}4.8$ & $104.5{\pm}10.3$ & -- \\
\midrule
\multirow{4}{*}{MMLU}
 & Recompute & $72.9{\pm}3.6$ & $94.2{\pm}2.1$ & -- \\
 & Stale     & $56.0{\pm}4.0$ & $23.9{\pm}0.3$ & 0 \\
 & \method   & $73.5{\pm}3.6$ & $24.4{\pm}0.4$ & 3.5 \\
 & KVCOMM    & $69.3{\pm}4.7$ & $84.5{\pm}4.7$ & -- \\
\midrule
\multirow{4}{*}{\shortstack[l]{Human-\\Eval}}
 & Recompute & $82.8{\pm}2.4$ & $60.9{\pm}8.5$ & -- \\
 & Stale     & $71.3{\pm}2.4$ & $27.3{\pm}0.4$ & 0 \\
 & \method   & $81.9{\pm}2.4$ & $30.7{\pm}0.3$ & 2.8 \\
 & KVCOMM    & $81.6{\pm}5.8$ & $78.2{\pm}0.2$ & -- \\
\bottomrule
\end{tabular}
\caption{Main results (harness protocol, A100). avg@4 is the mean correctness of one greedy and three $T{=}0.7$ samples, $\pm$ standard error across questions; paired differences are in Table~\ref{tab:paired}. TTFT is the per-agent median $\pm$ standard deviation of the foreground first-token latency; Table~\ref{tab:twostream} reports it with the exact prefill running concurrently. $K$ is the average optimistic window. GSM8K is all-optimistic; in MMLU and HumanEval the final agent uses the exact path.}
\label{tab:main}
\end{table}

\subsection{Quality and Time to First Token}

Table~\ref{tab:main} shows that permanent stale reuse obtains near-minimal \ttft but loses 11.5--16.9 accuracy points. RebaseKV keeps the low-latency path and recovers nearly all of this quality: $+0.6$ points relative to recomputation on MMLU, $-0.9$ on HumanEval, and $-2.6$ on GSM8K in the all-optimistic configuration, a gap the routing rule closes (Section~\ref{sec:paired}). The foreground \ttft falls from 79.3 to 26.3 ms on GSM8K ($3.0\times$), from 94.2 to 24.4 ms on MMLU ($3.9\times$), and from 60.9 to 30.7 ms on HumanEval ($2.0\times$), and the average draft is short, 2.8 to 3.7 tokens. Section~\ref{sec:twostream} confirms the GSM8K ratio with the exact prefill running concurrently.

\begin{table}[t]
\centering
\footnotesize
\setlength{\tabcolsep}{2.0pt}
\begin{tabular}{@{}lrrl@{}}
\toprule
Setting / metric & Rec. & Reb. & $\Delta$ [95\% CI] \\
\midrule
\multicolumn{4}{l}{\emph{GSM8K, Qwen2.5-7B: all-optimistic (A100, 1,318 q.)}} \\
\quad greedy & 91.5 & 88.2 & $-3.3$ [$-5.0$, $-1.7$] \\
\quad mean of 3, $T{=}0.7$ & 90.3 & 87.9 & $-2.4$ [$-3.4$, $-1.4$] \\
\quad avg@4 & 90.6 & 88.0 & $-2.6$ [$-3.5$, $-1.8$] \\
\multicolumn{4}{l}{\emph{GSM8K, Qwen2.5-7B: DM exact (A6000, 300 q.)}} \\
\quad greedy & 90.7 & 91.7 & $+1.0$ [$-1.0$, $+3.3$] \\
\quad avg@4 & 91.8 & 91.8 & $\phantom{+}0.0$ [$-1.2$, $+1.2$] \\
\multicolumn{4}{l}{\emph{GSM8K, Qwen3-8B, greedy (A6000, 300 q.)}} \\
\quad all-optimistic & 95.3 & 92.0 & $-3.3$ [$-6.7$, $-0.3$] \\
\quad DM exact & 95.3 & 94.3 & $-1.0$ [$-3.3$, $+1.3$] \\
\bottomrule
\end{tabular}
\caption{Paired accuracy (\%), RebaseKV (Reb.) minus Recompute (Rec.) on the same questions; recomputation is re-run on the same GPU for every A6000 block. DM: Decision Maker. Stale on GSM8K (A100): greedy 77.7, avg@4 75.4 ($-15.2$ [$-16.5$, $-13.9$]).}
\label{tab:paired}
\end{table}

\subsection{Paired Accuracy}
\label{sec:paired}

Table~\ref{tab:paired} reports the timed greedy sample separately from the sampled answers. Section~\ref{sec:format} traces the GSM8K all-optimistic gap ($-3.3$ greedy, $-2.4$ sampled) to a single answer-first agent. With that agent on the exact path, RebaseKV matches recomputation re-run on the same GPU and questions: avg@4 is 91.8 for both (paired difference 0.0, CI $-1.2$ to $+1.2$) and greedy accuracy is 91.7 vs.\ 90.7. Qwen3-8B shows the same pattern: the exact Decision Maker brings RebaseKV to within 1.0 point of recomputation, a difference not significant at 95\%.

\subsection{Measured Two-Stream Execution}
\label{sec:twostream}

\begin{table*}[t]
\centering
\small
\setlength{\tabcolsep}{4.6pt}
\begin{tabular}{lrrrrrrrr}
\toprule
 & Recompute & RebaseKV & & Exact prefill & & & & \\
Configuration & TTFT & TTFT & Speedup & alone $\to$ concurrent & $K$ & Stall & Replay & Overhead \\
\midrule
\multicolumn{9}{l}{\emph{Qwen2.5-7B-Instruct, 200 GSM8K questions}} \\
Fixed window & 135 (181) & \textbf{46} (58) & 2.95x & 137 $\to$ 235 & 6.5 & 0 & 28 & $+0.7\%$ \\
Event-driven window & 137 (185) & 46 (60) & 2.96x & 143 $\to$ 241 & 5.6 & 0 & 28 & $+0.4\%$ \\
Decision Maker exact & 134 (169) & 45 (57) & 2.94x & 136 $\to$ 228 & 6.3 & 0 & 28 & $+0.4\%$ \\
No stream priority & 137 (184) & 92 (97) & 1.49x & 143 $\to$ 182 & 6.5 & 0 & 27 & $+1.0\%$ \\
\midrule
\multicolumn{9}{l}{\emph{Qwen3-8B, 100 GSM8K questions}} \\
Fixed window & 159 (202) & \textbf{63} (84) & 2.54x & 160 $\to$ 238 & 5.6 & 0 & 37 & $-0.2\%$ \\
\bottomrule
\end{tabular}
\caption{Two-stream execution on one RTX A6000, greedy, with the foreground draft on a high-priority stream unless noted. All times are in ms, medians over the rebased agents (p90 in parentheses); the first agent and exact-path agents are excluded from TTFT. ``Exact prefill'' is the prefill of the new prompt alone vs.\ while the draft runs; ``stall'' is the wait for the fresh cache after the window; ``replay'' is the $K$-token forward on the fresh cache; ``overhead'' is the paired median change in non-decode time per five-agent pipeline (end-to-end time minus generated tokens $\times\,\tdecode$), relative to Recompute's end-to-end time. $\tdecode$ is 24 ms (Qwen2.5) and 33 ms (Qwen3).}
\label{tab:twostream}
\end{table*}

\begin{figure}[t]
    \centering
    \includegraphics[width=\columnwidth]{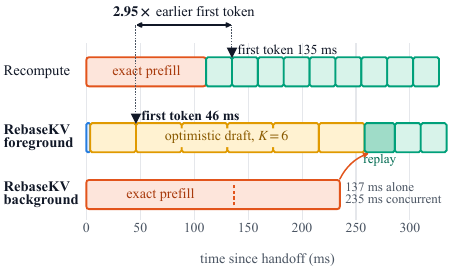}
    \caption{Measured handoff timeline (medians over the rebased agents, serving-style, Qwen2.5-7B, A6000). The first token arrives 2.95x earlier; the draft window outlasts the slowed exact prefill, so there is no stall; after replay both paths have emitted about the same number of tokens, so the first-token gain comes at no end-to-end cost.}
    \label{fig:measured}
\end{figure}

Table~\ref{tab:twostream} and Figure~\ref{fig:measured} run RebaseKV with the exact prefill actually executing concurrently with the optimistic draft. Four observations follow.

\emph{The first-token gain survives real concurrency.} In the serving-style configuration (Section~\ref{sec:execution}), the first token arrives in 46 ms instead of 135 ms (2.95x; p90 58 vs.\ 181 ms), matching the 3.0x of the harness protocol; Qwen3-8B obtains 2.54x.

\emph{Stream priority is the key implementation choice.} With the draft on a high-priority stream, contention is absorbed by the background prefill, which takes 235 instead of 137 ms, and by the draft steps, which take 41 instead of 24 ms, rather than by the first token; without priority the first draft step queues behind prefill kernels (1.49x).

\emph{No stall, one replay.} Because contended draft steps are slower, a window sized from the uncontended prefill still covers the slowed prefill: the stall is 0 ms at the median and at p90. The only added pause is the replay, a single $K$-token forward of 27--28 ms (37 ms for Qwen3-8B) per agent. The event-driven window gives the same first token with a shorter window ($K=5.6$).

\emph{Negligible end-to-end overhead.} RebaseKV moves exposed prefill off the first token at almost no cost: non-decode time per pipeline changes by $-0.2$ to $+1.0\%$, and WarmupKV, the exact pipelined alternative, behaves the same way ($+0.6\%$; Appendix~\ref{app:ablations}, which also reports a pipeline-depth ablation). Routing the Decision Maker to the exact path costs the other four handoffs nothing (2.94x).

\subsection{Comparison with RelayCaching}
\label{sec:relay}

\begin{table}[t]
\centering
\small
\setlength{\tabcolsep}{2.8pt}
\begin{tabular}{lrrr}
\toprule
Method & Acc. & TTFT, agents 2 / 3 / 4 / 5 & Speedup \\
\midrule
\multicolumn{4}{l}{\emph{RelayCaching harness}} \\
Full prefill & 90.0 & 95 / 128 / 170 / 228 & 1.00x \\
RelayCaching & 90.5 & 58 / 72 / 96 / 113 & 1.88x \\
\multicolumn{4}{l}{\emph{Our two-stream harness (Table~\ref{tab:twostream})}} \\
Recompute & 89.5 & 132 / 151 / 131 / 151 & 1.00x \\
RebaseKV & 89.5 & \textbf{45} / \textbf{45} / \textbf{46} / 152$^{\dagger}$ & \textbf{2.94x} \\
\bottomrule
\end{tabular}
\caption{RelayCaching vs.\ RebaseKV: Qwen2.5-7B-Instruct, the same five roles and first 200 GSM8K questions, greedy, batch size 1, one RTX A6000. TTFT is the per-agent median in ms; speedup is the median over reused (RelayCaching) or rebased (RebaseKV) agents. RelayCaching reuses 76\% of the KV entries. $^{\dagger}$Decision Maker on the exact path (Section~\ref{sec:routing}).}
\label{tab:relay}
\end{table}

We ran the public RelayCaching implementation with the same model, role set, questions, and GPU, using the layer range selected by its own profiler and its default thresholds (Appendix~\ref{app:relay}). Table~\ref{tab:relay} shows that RelayCaching preserves accuracy (90.5 vs.\ 90.0) and roughly halves TTFT, but its rectified prefill stays on the critical path and grows with the accumulated context, from 58 ms at agent 2 to 113 ms at agent 5. RebaseKV's first token costs one contended decode step at every rebased handoff (45--46 ms), independent of prompt length, and its long-lived cache is exact. The designs compose: a rectified cache is a better draft state, and the optimistic window can hide the rectification.

\section{Analysis}
\label{sec:analysis}

\subsection{The Optimistic Prefix Fixes the Format}
\label{sec:format}

\paragraph{Role adherence.}
The role probe isolates the failure mode of permanent reuse. Under recomputation, the Programming-Expert agent emits code in 100\% of GSM8K cases. Permanent stale reuse reduces this rate to 3\%, indicating that the inherited Analyst role continues to control generation. With a short optimistic window and replay, RebaseKV restores it to 98\%: replacing the long-lived stale state, rather than merely shifting its positions, is essential.

\paragraph{Where the residual gap comes from.}
Under recomputation the GSM8K Decision Maker answers in a median of 7 tokens. Under RebaseKV its first $K$ tokens are drafted from the Inspector's cache and open a step-by-step explanation, which the fresh cache then faithfully continues: the median length becomes 240 tokens and 6.4\% of pipelines hit the 400-token cap (0.2\% under recomputation). Of the 84 questions that only recomputation answers correctly under greedy decoding, 63\% are such truncations, and on the 1,231 questions without a Decision-Maker cap hit greedy accuracy is 92.1 for recomputation and 92.7 for RebaseKV. The optimistic prefix therefore changes the format of an answer-first agent rather than degrading reasoning.

\begin{table}[t]
\centering
\small
\setlength{\tabcolsep}{3.4pt}
\begin{tabular}{lrr}
\toprule
Policy & avg@4 (\%) & Code (\%) \\
\midrule
Recompute & 92.3 & 100 \\
Stale & 78.5 & 5 \\
Adaptive $K$ ($\bar K{=}3.7$) & 88.3 & 98 \\
$K=5$ & 88.2 & 99 \\
$K=10$ & 90.2 & 97 \\
$K=20$ & 89.9 & 98 \\
$K=50^{*}$ & 88.6 & 97 \\
\midrule
Adaptive $K$, DM exact$^{\dagger}$ & \textbf{91.8} & 97 \\
\bottomrule
\end{tabular}
\caption{Draft-length ablation on the first 300 GSM8K questions (Qwen2.5-7B). Code: Programming-Expert code-block rate. $^{*}$208 questions. $^{\dagger}$A6000 ($\bar K{=}7.9$); recomputation re-run on that GPU scores 91.8 (paired difference 0.0).}
\label{tab:k-ablation}
\end{table}

\paragraph{Routing, not window length, determines accuracy.}
Table~\ref{tab:k-ablation} varies $K$ on the first 300 GSM8K questions. Accuracy (88.2--90.2) and the role probe (97--99\%) are stable from $K=3.7$ to $K=50$, consistent with a format decision made in the first few tokens rather than an error that accumulates with the window. Routing the Decision Maker to the exact path (Section~\ref{sec:routing}) closes the gap even with the longer A6000 window ($\bar K=7.9$), removes the truncations (4.7\% to 0\% on these questions), and improves on the all-optimistic run by 3.4 points (95\% CI $+1.9$ to $+5.0$).

\subsection{Continuation Divergence}

Table~\ref{tab:kl-full} reports the cumulative continuation-KL ladder against full recomputation. Replaying immediately ($K=0$) isolates the bf16 execution-path residual, which reaches only 0.021 nats by token 60. RebaseKV with $K=3$ or $K=10$ remains close to this floor (0.09 and 0.11 nats). In contrast, a cache that is never rebased accumulates 3.46 nats. The result is consistent with Theorem~\ref{thm:finite-window}: after replay, the continuation state is fresh for the accepted prefix, whereas permanent stale decoding adds context-mismatch error at every step.

\begin{table}[t]
\centering
\small
\setlength{\tabcolsep}{2.6pt}
\begin{tabular}{lrrrrrr}
\toprule
Path & $t{=}1$ & $5$ & $10$ & $20$ & $40$ & $60$ \\
\midrule
Same cache twice & 0 & 0 & 0 & 0 & 0 & 0 \\
Stale, never rebased & 0.39 & 0.82 & 1.78 & 2.49 & 3.11 & 3.46 \\
Replay, $K=0$ & .001 & .002 & .004 & .008 & .014 & .021 \\
Rebase at $K=3$ & 0.01 & 0.01 & 0.02 & 0.04 & 0.06 & 0.09 \\
Rebase at $K=10$ & 0.01 & 0.02 & 0.03 & 0.05 & 0.09 & 0.11 \\
\bottomrule
\end{tabular}
\caption{Cumulative continuation divergence $\KL(\mathrm{fresh}\|\mathrm{path})$ in nats up to token $t$ (Qwen2.5-7B-Instruct, GSM8K). The $K=0$ replay path measures numerical differences from split bf16 execution; short windows remain near that residual, whereas permanent stale reuse accumulates mismatch.}
\label{tab:kl-full}
\end{table}

\section{Discussion}

\paragraph{When to use RebaseKV.}
RebaseKV targets consecutive agents that run the same model on prompts sharing most of their text, where the handoff latency is visible to a user and the next agent's first tokens are free-form. In our pipelines the changed role prefix is tens of tokens out of about 800. The system decides per agent from information it already has (Section~\ref{sec:routing}).

\paragraph{Resources.}
RebaseKV adds little GPU work. The exact prefill is computed under both recomputation and RebaseKV; the additions are the bootstrap rotation, $K$ draft steps that emit output tokens recomputation would also have to decode, and one $K$-token replay of roughly one decode step. The second cache lives only during the window (about 45 MB for an 800-token Qwen2.5-7B prompt). What RebaseKV uses is concurrency on one GPU for a few hundred milliseconds, which the high-priority draft absorbs without stalls while total pipeline time stays within 1\% (Table~\ref{tab:twostream}).

\paragraph{Relation to speculative decoding and rectification.}
RebaseKV needs neither a draft model nor a verification pass: the draft comes from a cache that already exists, and replay makes the long-lived state exact. A rectified draft state as in RelayCaching, confidence-aware early rebasing, and interference-aware schedulers can further improve the latency--quality trade-off within the same execution model.

\section{Conclusion}

We presented RebaseKV, a training-free method for hiding repeated prefill latency in prefix-divergent multi-agent LLM workflows. RebaseKV drafts a short prefix from an inherited cache while the exact prompt cache is computed concurrently, then replays the prefix on the fresh cache and continues standard decoding. With the exact prefill genuinely running concurrently on one GPU, it delivers the first token of each handoff 2.95x faster than recomputation (2.54x on Qwen3-8B), with no stall and negligible end-to-end overhead. The optimistic prefix mainly decides an agent's format; routing format-fixing agents to the exact path makes accuracy match recomputation while the other handoffs keep the low first-token latency. Role and divergence analyses show why the rebase boundary matters: it restores new-agent behavior and prevents stale cache states from governing the long continuation.

\section*{Limitations}\label{sec:limitations}

Our evaluation covers two Qwen2.5 7B variants and Qwen3-8B, three benchmark families, one multi-agent prompt design, and two GPU types; larger models, other positional encodings, tool-using agents, and workflows with less textual overlap are natural next steps. Agent routing is currently configured by agent type; driving it adaptively with the agreement monitor is a natural extension. Our concurrent measurements use batch size 1 on one GPU; a production-load study with batching and multi-tenant traffic is left for future work.

\section*{Ethical Considerations}

This work is a serving optimization and does not change the underlying model's training data, safety properties, or access controls. Lower latency may make multi-agent systems cheaper and more responsive, but it can also accelerate unreliable or harmful agent behavior if the underlying workflow is not adequately evaluated. RebaseKV is a latency optimization, not a safety mechanism. Our experiments use established public benchmarks and do not involve personal or sensitive data.

\bibliography{references}

\appendix
\section{Proofs}
\label{app:proofs}

\paragraph{Proof of Proposition~\ref{prop:cumulative-stale}.}
Equation~\ref{eq:stale-chain} decomposes sequence-level KL into the one-step terms $\delta_t$. The claim follows because KL divergence is nonnegative.

\paragraph{Proof of Theorem~\ref{thm:finite-window}.}
Apply the autoregressive KL chain rule to $\KL(P_{1:T}\|R^{(K)}_{1:T})$. For $t\le K$, the conditional comparison is $p_t$ versus $q_t$. For $t>K$, both sequence distributions use $p_t(\cdot\mid y_{<t})$, so every post-rebase term equals zero. Only the first $K$ stale-draft terms remain, yielding Equation~\ref{eq:finite-window}.

\paragraph{Proof of Corollary~\ref{cor:continuation}.}
Marginalization cannot increase KL divergence. Apply the data-processing inequality to the map that drops the first $K$ tokens, then use Theorem~\ref{thm:finite-window}.

\paragraph{Proof of Proposition~\ref{prop:logit-bound}.}
For a fixed history, write $q_t=\operatorname{softmax}((z_t+\Delta_t)/\tau)$. Then
\begin{equation}
\begin{aligned}
\KL(p_t\|q_t)
&=\log\mathbb{E}_{a\sim p_t}\!\left[
 e^{\Delta_t(a)/\tau}\right]\\[-2pt]
&\quad-\mathbb{E}_{a\sim p_t}\!\left[\Delta_t(a)/\tau\right].
\end{aligned}
\label{eq:hoeffding-step}
\end{equation}
Hoeffding's lemma bounds this quantity by $\operatorname{span}(\Delta_t)^2/(8\tau^2)$. Taking expectations and summing the first $K$ terms via Theorem~\ref{thm:finite-window} proves the claim.

\paragraph{Proof of Proposition~\ref{prop:greedy}.}
The fresh top-1 logit can decrease by at most $\|\tilde z_t-z_t\|_\infty$, while any competitor can increase by at most the same amount. The top-1 ordering is preserved when twice this perturbation is smaller than $\gamma_t$. Induction over the first $K$ steps establishes prefix identity; replay then reconstructs the same fresh state for that prefix.

\section{Detailed Transition Timeline}
\label{app:timeline}

\begin{figure*}[t]
    \centering
    \includegraphics[width=\textwidth]{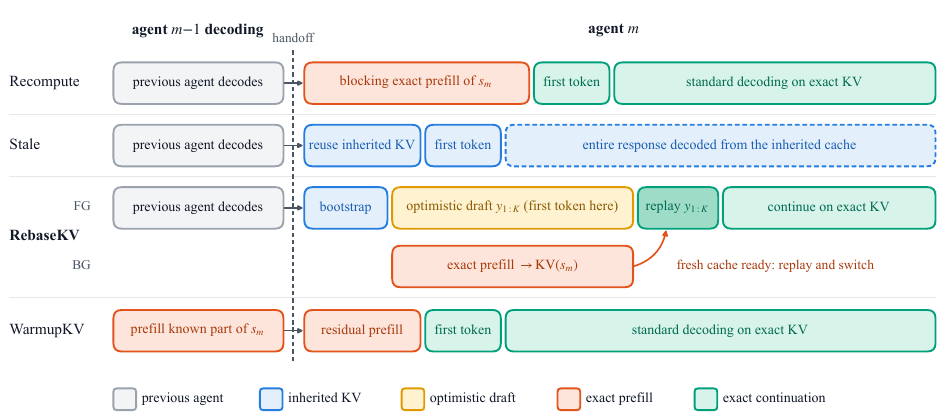}
    \caption{Schematic transition from agent $m-1$ to agent $m$ for every method. RebaseKV takes the exact prefill off the first-token critical path and switches to exact KV after one replay; WarmupKV hides the known part of the prefill during the previous agent's decoding. Figure~\ref{fig:measured} shows the measured counterpart.}
    \label{fig:timeline}
\end{figure*}

Figure~\ref{fig:timeline} expands the handoff schedule. Full recomputation blocks on exact prefill. Permanent stale reuse emits immediately but leaves the entire response role-misaligned. RebaseKV splits the handoff into a foreground draft and a background exact prefill, synchronizes once the fresh cache is ready, replays the accepted tokens, and then continues from prompt-consistent KV. WarmupKV starts exact work earlier, during the preceding agent's decode, but still exposes the residual produced by that agent.

\section{Two-Stream Execution Details}
\label{app:twostream}

\paragraph{Driver.}
For every question the driver runs a complete Recompute pipeline and a complete RebaseKV pipeline with the same model instance, greedy decoding, a 400-token cap per agent, and batch size 1. In the RebaseKV pipeline the first agent is exact. For each rebased agent the inherited cache is built outside the timed region (Section~\ref{sec:setup}); the exact prefill of the new prompt is then launched on a second CUDA stream and the foreground draft starts immediately. The serving-style configuration places the foreground on a high-priority stream and launches the background prefill from its own CPU thread; the naive configuration uses the default stream and launches the prefill from the foreground thread. Draft steps, replay, and the background prefill are timed with CUDA events; TTFT is wall-clock time from the start of the handoff to the first drafted token. We also time the exact prefill of every new prompt alone to obtain the uncontended reference and $K$.

\paragraph{Per-agent breakdown.}
Table~\ref{tab:per-agent} lists per-agent medians for the serving-style fixed-$K$ configuration. TTFT is flat across rebased agents because it is one contended decode step, whereas recomputation TTFT follows the prompt length.

\begin{table}[h]
\centering
\small
\setlength{\tabcolsep}{5pt}
\begin{tabular}{@{}lrrrrr@{}}
\toprule
 & \multicolumn{2}{c}{TTFT} & & Prefill & \\
\cmidrule(lr){2-3}
Agent & Rec. & Reb. & $K$ & conc. & Replay \\
\midrule
Math Solver & 196 & 196 & -- & -- & -- \\
Analyst & 132 & 45 & 6 & 224 & 27.5 \\
Prog.\ Expert & 152 & 45 & 7 & 239 & 27.6 \\
Inspector & 131 & 46 & 6 & 233 & 27.8 \\
Decision Maker & 151 & 46 & 7 & 251 & 27.9 \\
\bottomrule
\end{tabular}
\caption{Per-agent medians (ms) of the serving-style, fixed-$K$ two-stream run (Qwen2.5-7B, 200 GSM8K questions, RTX A6000). ``Prefill conc.'' is the exact prefill while the draft runs; the first agent (Math Solver) has no inherited cache and runs the exact path.}
\label{tab:per-agent}
\end{table}

\section{RelayCaching Setup}
\label{app:relay}

We used the public RelayCaching implementation with Qwen2.5-7B-Instruct and its five-agent GSM8K setting with the KVCOMM role set, the first 200 test questions, greedy decoding, and batch size 1 on the same RTX A6000 as Section~\ref{sec:twostream}. Its layer profiler selected start layer 2, detection layer 3, and end layer 25 for this model; all thresholds were left at their defaults. The released GSM8K loader omitted appending the parsed items to its dataset list, so every mode exited immediately with an empty dataset; we added the missing append and changed nothing else. Accuracy and TTFT are as reported by its harness, where TTFT is the wall time to the first generated token inside its generation call. Its prompts contain no few-shot exemplars, which is one reason the full-prefill accuracy differs from our harness (90.0 vs.\ 89.5).

\section{System Ablations}
\label{app:ablations}

Table~\ref{tab:system-ablations} reports two secondary system ablations on the A100 harness. First, WarmupKV asks how much prefill can be hidden without optimistic decoding: it prefills the known portion of agent $m$ while agent $m-1$ is decoding, then processes only the newly produced residual at handoff. It matches full-recompute task accuracy on all three benchmarks, but the residual remains on the \ttft path, yielding 35.1--63.1 ms versus 24.2--30.2 ms for RebaseKV in the corresponding runs. ``Exact'' here means prompt-consistent computation; split bf16 forwards can differ at the bit level because their accumulation order changes. In a real two-stream GSM8K run (1,319 questions), WarmupKV lowers the per-agent TTFT from 78.5 to 63.1 ms while the end-to-end time changes from 21.37 to 21.56 s (paired median ratio 1.006), the same under-1\% behavior as RebaseKV.

Second, we vary the number of HumanEval worker agents from two to five, always followed by a Decision Maker. RebaseKV shows no monotonic accuracy degradation, median \ttft stays near one decode step, and the average draft remains 2.04--2.46 tokens. The ideal-overlap hidden prefill grows from 16.3 to 92.5 ms as handoffs are added, which is 0.44--1.71\% of the local prefill-plus-draft region because response decoding dominates the full round.

\begin{table*}[t]
\centering
\footnotesize
\setlength{\tabcolsep}{4.1pt}
\begin{tabular}{lrrrrr}
\toprule
\multicolumn{6}{l}{\textbf{(a) Exact-pipelining ablation: TTFT and ideal-overlap hidden prefill (ms)}} \\
Dataset & Recompute TTFT & WarmupKV TTFT & RebaseKV TTFT & WarmupKV hidden & RebaseKV hidden \\
\midrule
GSM8K & $79.3{\pm}12.3$ & $63.1{\pm}11.5$ & $26.3{\pm}0.2$ & $73{\pm}16$ & 200 \\
MMLU & $94.7{\pm}14.8$ & $46.6{\pm}5.8$ & $24.2{\pm}0.4$ & $294{\pm}110$ & 420 \\
HumanEval & $60.4{\pm}16.7$ & $35.1{\pm}4.3$ & $30.2{\pm}0.3$ & $114{\pm}57$ & 269 \\
\midrule
\multicolumn{6}{l}{\textbf{(b) Pipeline-depth ablation on HumanEval}} \\
Workers $N$ & Recompute acc. (\%) & RebaseKV acc. (\%) & RebaseKV TTFT (ms) & Avg. $K$ & Hidden ms (ratio) \\
\midrule
2 & $83.4{\pm}2.9$ & $84.2{\pm}2.9$ & $31.1{\pm}11.6$ & 2.04 & 16.3 (0.44\%) \\
3 & $81.7{\pm}3.0$ & $85.9{\pm}2.7$ & $31.1{\pm}14.8$ & 2.10 & 36.9 (0.86\%) \\
4 & $80.6{\pm}3.1$ & $83.4{\pm}2.9$ & $31.6{\pm}23.0$ & 2.24 & 85.8 (1.39\%) \\
5 & $82.8{\pm}3.0$ & $81.5{\pm}3.1$& $30.2{\pm}30.0$ & 2.46 & 92.5 (1.71\%) \\
\bottomrule
\end{tabular}
\caption{Secondary system ablations on a dedicated A100 40GB (harness protocol). In (a), TTFT is median $\pm$ standard deviation, and hidden prefill per pipeline assumes ideal overlap (Section~\ref{sec:twostream} measures real overlap); WarmupKV matches recomputation accuracy. In (b), accuracy is avg@4 $\pm$ standard error, TTFT is median $\pm$ standard deviation, and workers are added in a fixed role order before the final Decision Maker.}
\label{tab:system-ablations}
\end{table*}

\end{document}